\documentclass[10pt,twocolumn,letterpaper]{article}

\usepackage{iccv}              

\usepackage{makecell}
\usepackage{amssymb}
\usepackage{xcolor}
\definecolor{darkgreen}{RGB}{0,100,0}
\definecolor{mediumgreen}{RGB}{0,128,0}
\definecolor{lightgreen}{RGB}{144,238,144}
\definecolor{limegreen}{RGB}{50,205,50}
\definecolor{forestgreen}{RGB}{34,139,34}
\newcommand{\ca}[1]{{\color{red} \bf \em (#1)}}

\newcommand{\ours}{GS-Marker}

\definecolor{iccvblue}{rgb}{0.21,0.49,0.74}
\usepackage[pagebackref,breaklinks,colorlinks,allcolors=iccvblue]{hyperref}
\usepackage{graphicx}
\usepackage{booktabs}
\usepackage{multirow}
\usepackage{amssymb}  
\usepackage{pifont}  
\usepackage{adjustbox}  
\usepackage{array}  
\usepackage{caption}
\usepackage{tabularx}  

\def\paperID{9482} 
\def\confName{ICCV}
\def\confYear{2025}

\title{\ours: Generalizable and Robust Watermarking for 3D Gaussian Splatting}

\author{Lijiang Li\thanks{This work was done during Lijiang Li's internship at MSRA.}, ~~ Jinglu Wang$^{1}$, ~~ Xiang Ming$^{1}$, ~~ Yan Lu$^{1}$ \\
Microsoft Research Asia$^{1}$\\
{\tt\small lijiangli.cs@gmail.com, \{jinglwa, xiangming, yanlu\}@microsoft.com}
}

\begin{document}

\maketitle

\begin{abstract}

In the Generative AI era, safeguarding 3D models has become increasingly urgent. While invisible watermarking is well-established for 2D images with encoder-decoder frameworks, generalizable and robust solutions for 3D remain elusive. The main difficulty arises from the renderer between the 3D encoder and 2D decoder, which disrupts direct gradient flow and complicates training. Existing 3D methods typically rely on per-scene iterative optimization, resulting in time inefficiency and limited generalization. In this work, we propose a single-pass watermarking approach for 3D Gaussian Splatting (3DGS), a well-known yet underexplored representation for watermarking. We identify two major challenges: (1) ensuring effective training generalized across diverse 3D models, and (2) reliably extracting watermarks from free-view renderings, even under distortions. Our framework, named \ours, incorporates a 3D encoder to embed messages, distortion layers to enhance resilience against various distortions, and a 2D decoder to extract watermarks from renderings. A key innovation is the Adaptive Marker Control mechanism that adaptively perturbs the initially optimized 3DGS, escaping local minima and improving both training stability and convergence. Extensive experiments show that \ours~outperforms per-scene training approaches in terms of decoding accuracy and model fidelity, while also significantly reducing computation time.

\end{abstract}



\section{Introduction}

\begin{figure}
  \centering
   \includegraphics[width=1\linewidth]{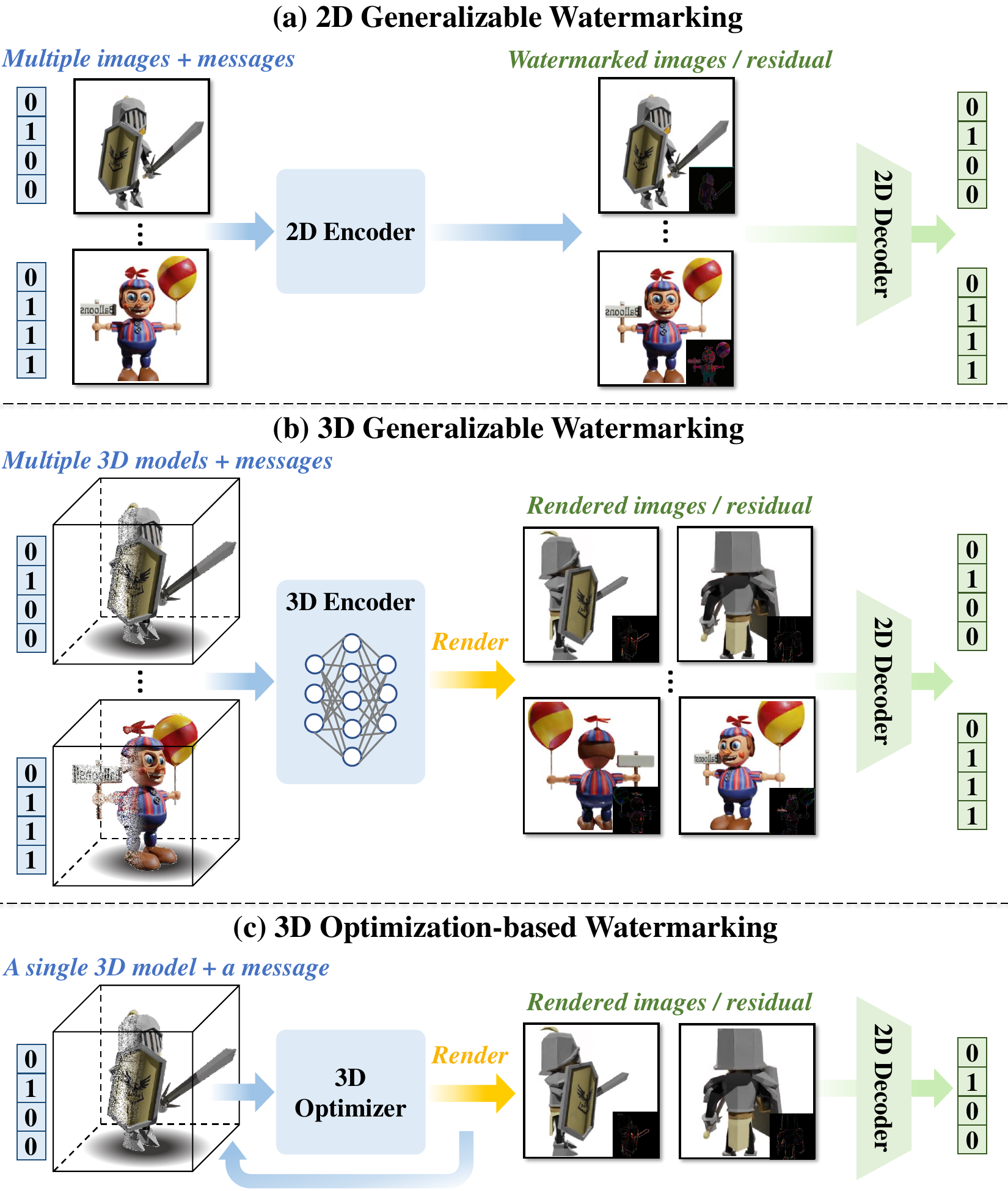}
   \caption{
(a) Generalizable 2D watermarking handles multiple images and messages in a single forward pass. (b) Extending to 3D is challenging due to the renderer bottleneck that disrupts direct gradient flow between the encoder and decoder. (c) Existing 3D approaches \cite{copyrnerf,waterf,gs_hider,NEURIPS2024_39cee562,jang20243d} often rely on iterative optimization for each model and message, which is time-consuming. In this paper, we tackle generalizable 3D watermarking, enabling single-pass embedding of watermarks into 3D models and robust extraction from their rendered images.
   }
  \label{fig:teaser1}
\end{figure}

With the rise of AI-generated assets, media provenance is more critical than ever. Invisible watermarking, which embeds watermarks without degrading quality, has emerged as a key solution. 2D methods \cite{hidden} for images, use an encoder-decoder framework for efficient, generalizable message embedding and extraction, as shown in \cref{fig:teaser1} (a).

Extending generalizable 2D watermarking, which embeds watermark in a single-forward pass, to 3D is challenging. Watermarks must be incorporated into the 3D models while remaining extractable from 2D renderings, as illustrated in \cref{fig:teaser1} (b).
The differentiable renderer between the encoder and decoder often becomes a gradient bottleneck, complicating end-to-end training.
Most existing 3D methods \cite{copyrnerf,waterf,gs_hider,NEURIPS2024_39cee562,jang20243d} rely on multi-iteration optimization for each model and message (\cref{fig:teaser1} (c)), limiting their efficiency and generalization. For example, CopyNeRF \cite{copyrnerf} and WaterRF \cite{waterf} require hours per NeRF \cite{nerf} model. Recent 3D Gaussian Splatting (3DGS) methods are all optimization-based, the fastest GaussianMarker \cite{NEURIPS2024_39cee562} needs about six minutes per model. Research on generalizable 3D watermarking for diverse 3DGS models is limited, hindered by the challenges of unstructured 3D data and differentiable rendering. As 3DGS gains prominence, the demand for efficient, scalable solutions grows. Although Deep-3D-to-2D \cite{mesh_watermark1} generalizes for meshes, its mesh-specific encoding and rendering processes are incompatible with 3DGS.

In this paper, we propose a novel framework, \ours, for robust and generalizable watermarking of 3D Gaussian Splatting (3DGS) in a single forward pass. We address two key challenges: (1) developing a 3D encoder that generalizes across diverse 3DGS models, and (2) ensuring accurate watermark extraction from free-view renderings under various distortions. 

\ours~comprises a 3D encoder that embeds messages within 3DGS and a 2D decoder that extracts them from rendered images. Both components are jointly optimized to balance watermark decoding accuracy and rendering fidelity. 
However, because the original 3DGS is already optimized for rendering, it lies near a local minimum for fidelity, making direct watermark embedding prone to large, destabilizing updates. To address this, we introduce an Adaptive Marker Control (AMC) mechanism that adaptively perturbs each 3DGS property, shifting it away from the local optimum. Specifically, we sample noise for each property and scale it with a learnable weight, so the perturbation magnitude adapts to the property’s sensitivity. As a result, watermark and rendering objectives can be balanced, guiding training more effectively.
Then, We feed the perturbed 3DGS into a point-based embedding network, which predicts residual updates for each primitive property. Additionally, \ours~is designed to withstand a variety of distortions, allowing the decoder to recover messages even from degraded renderings.

In summary, our contribution is three-fold.
\begin{itemize}
\item 
A generalizable, robust watermarking solution for 3DGS that is the first to require no per-scene training, significantly reducing computational overhead.

\item 
An adaptive marker control mechanism to avoid local minima and balance watermark accuracy with rendering fidelity.
\item
Comprehensive experiments demonstrating superior performance in both accuracy and fidelity over existing methods requiring per-scene optimization.
\end{itemize}

\section{Related Works}
\label{sec.related_work}
\paragraph{3D Gaussian splatting.}
3D Gaussian splatting (3DGS) is an efficient and powerful technique for reconstructing and representing 3D models. Specifically, 3DGS employs a set of Gaussian ellipsoids to approximate the geometry and appearance of a scene or object, and utilizes the differentiable tile rasterizer \cite{3dgs} for rendering.
This method enables fast and high-quality 3D reconstruction for complex scenes and objects, making 3DGS applicable in a wide range of domains, such as scene reconstruction \cite{3dgs_scene1,3dgs_scene2,3dgs_scene3}, objects generation \cite{LGM,triplaneGaussian,GRM}, and virtual human. \cite{3dgs_avatars1,3dgs_avatars2,3dgs_avatars3,3dgs_avatars4,3dgs_avatars5}. 
Given the rapid developments and increasing adoption of 3DGS in recent years, the protection of intellectual property in this domain has become a critical concern. Therefore, we propose a robust watermarking framework for 3DGS to address this issues in this work.

\paragraph{2D Digital watermarking.}
Digital watermarking is a crucial technique for ensuring the security of digital assets. This technique typically embeds a watermark message into digital content while preserving the perceptual quality of the watermarked product. The embedded watermark must be retrievable from the watermarked content without significant information loss. A seminal works in this field is HiDDeN \cite{hidden}, which introduced a encoder-decoder pipeline for image watermarking. Following this pioneering work, many subsequent studies have extended watermarking techniques into images \cite{image_watermark1,image_watermark2,image_watermark3,image_watermark4,image_watermark5} and videos \cite{video_watermark1,video_watermark2,video_watermark3}.

\paragraph{3D Digital watermarking.}
In the realm of 3D content, numerous existing works have introduced robust watermarking approaches for 3D models with various representation. For mesh and point cloud models, the mainstream approach usually involves embedding the watermark message on the vertex and points \cite{mesh_watermark1,mesh_watermark2,mesh_watermark3,point_watermark}. For instance, \cite{mesh_watermark1} proposed a deep watermarking pipeline for 3D meshes, which embeds messages into the vertex and texture components of the input meshes. For NeRF, a typical watermarking method involves integrating the watermark message into the neural field through fine-tuning the NeRF models \cite{nerf_watermark1,copyrnerf,waterf}. For 3DGS, GaussianMarker \cite{NEURIPS2024_39cee562} and 3D-GSW \cite{jang20243d} are pioneers of watermarking 3DGS, which succeed in embedding watermark messages into the 3DGS models through optimization. Although these methods have demonstrated impressive performance on 3DGS watermarking, they require extensive fine-tuning for embedding messages when a new model comes, which may limit their application in real-world scenarios. In this work, we aim to propose a 3D watermark framework specifically designed for 3DGS. Distinct from existing 3DGS watermarking methods, our framework enables watermark embedding in a single forward, which is more efficient and generalizable. 

{
\section{Method}

\begin{figure}[t!]
    \centering
    \includegraphics[width=0.9\linewidth]{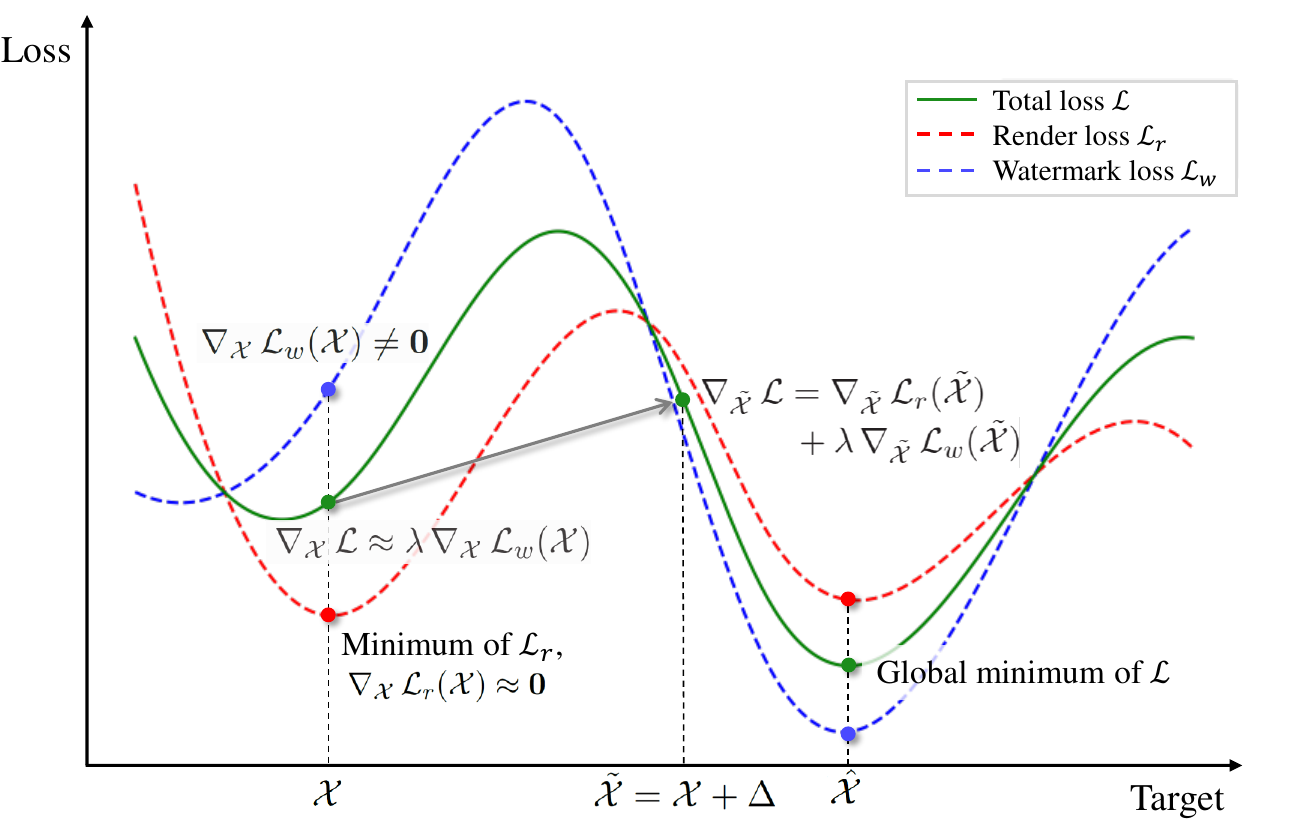}
    \caption{\textbf{Escape local minimum}. Since the input $\mathcal{X}$ lies near a local minimum of the rendering loss $\mathcal{L}_r$, its gradient is near 0. The total loss gradient $\nabla_{{\mathcal{X}}}\,\mathcal{L} \approx \lambda\,\nabla_{{\mathcal{X}}}\,\mathcal{L}_w({\mathcal{X}})$, which can conflict with $\mathcal{L}_r$. To address this, we introduce a suitable perturbation $\boldsymbol{\Delta}$ for $\mathcal{X}$, helping the model escape the local minimum and balance both objectives effectively.}
    \label{fig:min_illustration}
\end{figure}
\paragraph{Problem setup.}
We begin with a 3D Gaussian Splatting representation 
$\mathcal{X} \in \mathbb{R}^{N \times d}$,
where \(N\) is the number of Gaussian primitives, 
and each primitive has \(d\) properties 
(e.g., color, opacity, position, scale and orientation). The objective is to embed a binary message $M \in \{0, 1\}^{l_b}$ into 3DGS such that it can be accurately extracted from any rendered images, while preserving high rendering fidelity. Formally, we seek an optimized 3DGS $\hat{\mathcal{X}}$ by minimizing a combined loss consisting of watermark decoding and rendering terms
\begin{align}
 \hat{\mathcal{X}}
=
\arg\min_{\hat{\mathcal{X}}}
\Bigl(
  \mathcal{L}_w(\hat{\mathcal{X}})
  + 
  \lambda\,\mathcal{L}_r(\hat{\mathcal{X}})
\Bigr),   
\end{align}
where $\lambda$ is a weighting factor that trades off between watermarking accuracy and rendering fidelity.

The watermark loss $\mathcal{L}_w$ measures the accuracy of extracting embedded messages from rendered images. Let $I(\hat{\mathcal{X}}, v)$ denote the image rendered from  $\hat{\mathcal{X}}$ at viewpoint $v$,  $M_{pred}$ the predicted message from the rendered image. $\mathcal{L}_w$ is then defined as the mean squared error (MSE) between the original and predicted messages:
\begin{equation}
\mathcal{L}_w(\hat{\mathcal{X}})=\mathcal{L}_{mse}(M, M_{pred}(I(\hat{\mathcal{X}}, v))),
\end{equation}

The rendering loss $\mathcal{L}_r$ ensures that the rendered images $I(\hat{\mathcal{X}}, v)$ from the watermarked 3DGS $\hat{\mathcal{X}}$ closely match those $I(\mathcal{X}, v))$ from the original 3DGS $\mathcal{X}$ at any viewpoint $v$, thereby preserving visual fidelity.
\begin{equation}
\resizebox{.9\hsize}{!}{$\mathcal{L}_r(\hat{\mathcal{X}}) =
    \mathcal{L}_{mse}(I(\hat{\mathcal{X}}, v), I(\mathcal{X}, v)) + 
    \mathcal{L}_{lpips}(I(\hat{\mathcal{X}}, v), I(\mathcal{X}, v))$}
\end{equation}
where $L_{mse}$ ensures numerical similarity and $L_{lpips}$ measures perceptual similarity based on LPIPS~\cite{lpips}.

\subsection{Adaptive Marker Control}
\label{sec:amc}

\paragraph{Local minima challenge.}
The original 3DGS $\mathcal{X}$ is well optimized for rendering ($\mathcal{L}_r$ alone). \cref{fig:min_illustration} illustrates the situation.
Thus, $\mathcal{X}$ is near a local minimum for the rendering objective, leading to small gradients (at flat spot) for $\mathcal{L}_r$.
Meanwhile, adding the watermark objective $\mathcal{L}_w$ can produce large gradients dominate the total loss, which may conflict with $\mathcal{L}_r$. 


\[
\nabla_{{\mathcal{X}}}\,\mathcal{L}_r({\mathcal{X}})
\approx
\mathbf{0},
\quad
\text{but}
\quad
\nabla_{{\mathcal{X}}}\,\mathcal{L}_w({\mathcal{X}})
\neq \mathbf{0}.
\]
\[
\Longrightarrow
\quad
\nabla_{{\mathcal{X}}}\,\mathcal{L}
\approx
\lambda\,\nabla_{{\mathcal{X}}}\,\mathcal{L}_w({\mathcal{X}}).
\]


As a result, the network might struggle to find a meaningful descent direction, or it might cause large initial updates that derail fidelity. The optimization often fails to balance the two losses.
This mismatch can cause instability or suboptimal convergence if we simply refine $\mathcal{X}$ directly.


\paragraph{Adaptive perturbation.}
To escape from the local minimum, we propose to perturb $\mathcal{X}$ first. 
By adding $\boldsymbol{\Delta}$ to form $\tilde{\mathcal{X}}=\mathcal{X}+\boldsymbol{\Delta}$, we effectively move out of the local optimum for $\mathcal{L}_r$. This changes the gradient geometry. $\tilde{\mathcal{X}}$ is not strongly locked into the local basin of $\mathcal{L}_r$. This ensures that both $\mathcal{L}_r$ and $\mathcal{L}_w$ provide meaningful gradients right from the start, improving convergence to a solution that balances both objectives.

As the input 3DGS $\mathcal{X}$ is often highly dense, perturbing a subset of primitives is sufficient for watermark embedding while reducing training complexity. The subset size is a trade-off between embedding capacity and training stability, requiring careful selection.
Furthermore, different 3DGS properties require different perturbation magnitudes. Intuitively, highly sensitive properties (e.g., position, where small changes can significantly impact rendering) should have minimal perturbation, while properties with more flexibility for watermark embedding can tolerate larger perturbations.

To address this, we introduce an \textbf{Adaptive Marker Control (AMC)} mechanism. 
The perturbation $\boldsymbol{\Delta}(\boldsymbol{\omega})$ is modulated by learnable amplitudes $\omega_{j}$ of each property dimension $j$ and selection masks $m_i \in \{0,1\}$, where $m_i=1$ indicates that the $i$-th Gaussian primitive is selected for embedding, otherwise $m_i=0$. The perturbed 3DGS is formulated as
\begin{equation}
\tilde{\mathcal{X}}
=
\mathcal{X}
+
\boldsymbol{\Delta}(\boldsymbol{\omega}), \quad
\text{where}
\quad
\Delta_{i,j}(\omega)
=
\omega_{j}\,P_{i,j} m_i.
\end{equation}

$P_{i,j}$ denotes the noise sample for the $j$-th property of the $i$-th primitive draw from a Gaussian distribution, $P_{i,j} \sim \mathcal{N}\bigl(0,\;1\bigr)$. 
The mask $m_i$ is a random chosen based on a predefined selection ratio $\gamma \in[0,1]$, i.e., $m_i \sim \text{Bernoulli}(\gamma)$.
This adaptive strategy enhances both training stability and embedding effectiveness.

\subsection{Network Design}
\label{sec:network}

\begin{figure*}[t!]
  \centering
   \includegraphics[width=1\linewidth]{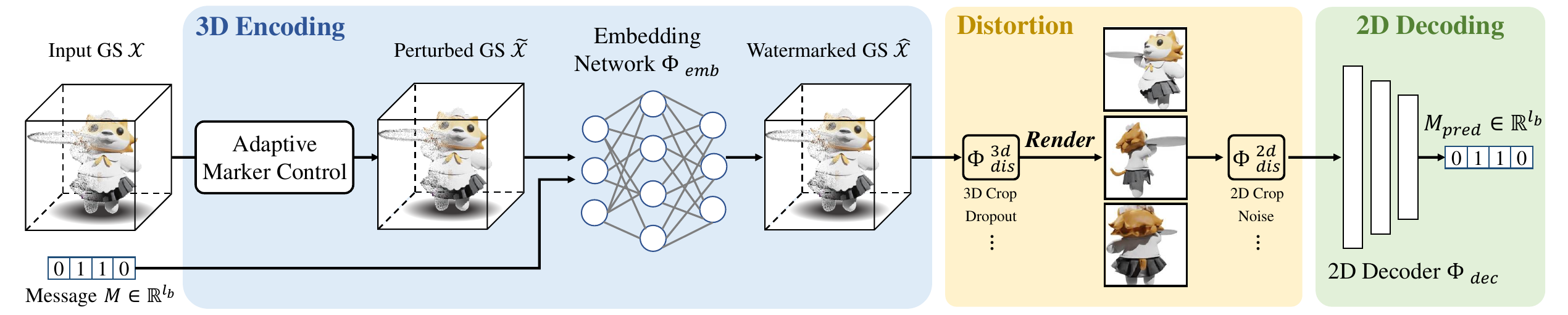}
   \caption{
   \textbf{Method overview.} 
   Given the input 3DGS $\mathcal{X}$ and the watermark $M$, \ours~embeds $M$ into $\mathcal{X}$ within a single forward pass, allowing $M$ to be extracted from rendered images even after undergoing both 3D and 2D distortions. Our framework comprises three key components: (1) a \textbf{3D encoding module}, which applies adaptive marker control to adjust $\mathcal{X}$, forming $\tilde{\mathcal{X}}$ to escape local minima, followed by an embedding network that transforms $(\tilde{\mathcal{X}}, M)$ into the watermaked 3DGS $\hat{\mathcal{X}}$. (2) a \textbf{distortion module}, incorporating 3D and 2D distortions to enhance robustness; and (3) a \textbf{2D decoding} module, which extracts the watermark from rendered images.
   }
  \label{fig:pipeline}
\vspace{-10pt}
\end{figure*}

\cref{fig:pipeline} illustrates our framework, which consists of three key modules: (1) an embedding network $\Phi_{emb}$ that integrate the watermark $M$ into the perturbed 3DGS $\tilde{\mathcal{X}}$; (2) distortion layers $\Phi_{dis}$ that enhance robustness against various degradations; (3) a decoder $\Phi_{dec}$ that extracts the embedded message from the rendered images. 

\paragraph{Embedding network.}
We employ an embedding network $\Phi_{emb}$ to integrate watermark $M$ into the perturbed 3DGS $\tilde{\mathcal{X}}$.
Follow common practices in 2D watermarking, our approach predicts the residual adjustiment to $\tilde{\mathcal{X}}$, ensuring minimal yet effective modifications while preserving rendering fidelity. 
\begin{equation}
\hat{\mathcal{X}}=\tilde{\mathcal{X}}+\Phi_{emb}\bigl(\tilde{\mathcal{X}}, M\bigr).
\end{equation}

As the input 3DGS consists of unordered Gaussian primitives, we adopt a point-based network $\Phi_{emb}$ inspired by \cite{pointnet++}, to ensure generalization.
Specifically, we first normalize all Gaussian properties into the range $[-1, 1]$, forming a concatenated vector $\mathbf{x}_i=[\boldsymbol{\mu}_i, \boldsymbol{\alpha}_i, \mathbf{c}_i, \mathbf{r}_i, \mathbf{s}_i] \in \mathbb{R}^{d_g}$, where $\boldsymbol{\mu}_i$ is the position, $\boldsymbol{\alpha}_i$ is the opacity, $\mathbf{c}_i$ is the color, $\mathbf{r}_i$ is the rotation,  $\mathbf{s}_i$ is the scaling for the $i$-th Gaussian. $\mathbf{x}_i$ is then perturbed by the adaptive marker control to form a modified feature $\tilde{\mathbf{x}}_i$. Each $\tilde{\mathbf{x}}_i$ is concatenated with the watermark message $M$, forming $\tilde{\mathbf{x}}_i^M = [\tilde{\boldsymbol{\mu}}_i, \tilde{\boldsymbol{\alpha}}_i, \tilde{\mathbf{c}}_i, \tilde{\mathbf{r}}_i, \tilde{\mathbf{s}}_i, M] \in \mathbb{R}^{(d_g+l_b)}$,
which serves as the input for $\Phi_{emb}$. $\Phi_{emb}$ consists of multiple cascaded blocks. Each block groups primitives into local regions based on their center positions. The features of primitives within the same region are aggregated using a pooling layer and subsequently refined through an MLP, generating the block’s output. The final output of $\Phi_{emb}$ is a residual adjustment $\boldsymbol{\delta} \in \mathbb{R}^{N \times d_g}$, which is added to the perturbed $\tilde{\mathcal{X}}$ 
 , yielding the final watermarked 3DGS $\hat{\mathcal{X}} = \tilde{\mathcal{X}} + \boldsymbol{\delta}$.

\paragraph{Distortion layer.}
\label{sec.method_distortion_layer}
To ensure robustness, the optimized 3DGS $\hat{\mathcal{X}}$ must be resilient to both 3D distortions applied directly to the 3DGS and 2D distortions applied to its rendered images.
The distorted image is formulated as:
\[
\hat{I}_{dis} = \Phi_{dis}^{2d}(I(\Phi_{dis}^{3d}(\hat{\mathcal{X}}), v))
\]
where $\Phi_{dis}^{3d}$ represents 3D distortions, chosen from \{primitive dropout, 3D cropping\} and $\Phi_{dis}^{2d}$ represents 2D distortions, chosen from \{noise, JPEG compression, cropping, rotation\}. The distortions are differentiable, ensuring compatibility with end-to-end training.



\paragraph{2D watermark decoder.}
\label{sec.method_2D_decoder}
We employ a 2D decoder $\Phi_{dec}$ to extract the embedded message from the distorted rendered image $\hat{I}_{dis}$. Specifically, we utilize the HiDDeN decoder \cite{hidden} for its widespread use. It applies multiple 2D convolution layers for feature extraction, followed by global spatial pooling to generate the predicted message $M_{pred}$. 

} 
\section{Experiments}
\label{sec.exps}
\subsection{Experimental Setting}
\paragraph{Dataset.}
To validate the generalizability of our method, we use datasets from both the source domain (for training and testing) and the target domain (for testing only).

\begin{itemize}
    \item For the \textbf{source domain}, we adopt the widely used Objaverse dataset \cite{objaverse}.
Specifically, we randomly sampled 1500 objects. For each object, we render 80 views using the official rendering scripts \cite{objaverse}.
We then optimize the original 3DGS using these rendered views with the official Gaussian splatting codebase \cite{3dgs}, forming the Objaverse$^{GS}_{train}$ dataset.
In addition, we sample another 100 objects from Objaverse to create a source-domain test set, ObjaverseGS$^{GS}_{test}$.

\item For the \textbf{target domain}, we follow standard practice in object-based 3DGS studies, and employ the Blender \cite{nerf} and OmniObject3D \cite{wu2023omniobject3d} datasets for further evaluation. Since initializing the original 3DGS is resource-intensive, we randomly sample 100 objects from OmniObject3D and construct original 3DGS to form a test set, Omniobject3D$^{GS}$. For Blender, we use all objects to form a test set, Blender$^{GS}$. 
\end{itemize}


\paragraph{Metrics.} The performance of our method is evaluated in terms of both the visual quality and data capacity. For visual quality, we evaluate the PSNR, SSIM, and LPIPS \cite{lpips} between the images rendered by the original and watermarked 3DGS. For data capacity, we calculate the bit accuracy of the extracted message from images rendered from watermarked 3DGS under various lengths of message $N_b \in \{4, 8, 16, 32, 64, 128\}$. In all experiments, we evaluate our methods as well as the baseline in our 3DGS test dataset with 100 objects. For each object in the test dataset, we uniformly sample 40 camera poses around the input 3D object to calculate the evaluation matrices. These camera poses used in the evaluation process have elevation sampled uniformly from $[-5^\circ, 5^\circ]$ and azimuth with value $9^\circ \times r, r\in [0, 39]$.  

\paragraph{Implementation details.}

The training is conducted on 8 V100 GPUs.
We first pre-train the HiDDeN \cite{hidden} model on $20,000$ sampled images from the rendered images of the Objaverse dataset and utilize the pretrained HiDDeN decoder to initialize our 2D decoder.
For training \ours, we utilize the Adam optimizer with a learning rate $1\times 10^{-4}$ for both the encoder and decoder. Due to the varying number of Gaussian primitives across different 3DGS models, the batch size is set to 1 in the training. In each iteration, images are rendered from 4 randomly sampled camera poses for both the watermarked and input 3DGS, with the elevation and azimuth of the camera poses sampled from the intervals $[-90^\circ, 90^\circ]$ and $[0^\circ, 360^\circ]$, respectively. The models are trained for $800,000$ iterations. We divide the training into three stages to implement our adaptive weighting strategy. The first stage spans from iteration $0$ to $200,000$, the second from iteration $200,000$ to $500,000$, and the final stage from iteration $500,000$ to $800,000$. Initially, the training is conducted without applying any distortion. Upon completion of the three training stages, we add the distortion layer and continue training for an additional $400,000$ iterations. Our experiments are conducted on message length $l_b=16$ unless mentioned specifically.

The learnable perturbation amplitude $\omega_j$ for Gaussian properties are initialized as follows: position at 0.01, scaling at 0.5, rotation at 0.1, and both opacity and color at 1.0. The selection ratio $\gamma$ is chosen as 0.4 and will be examined in detail in \cref{sec:ablation}.

\paragraph{Baseline.} Given the absence of generalizable watermarking approaches for 3DGS, we compare our method against existing 2D and optimization-based 3D watermarking strategies.
For the 2D watermarking strategy, we incorporate the HiDDeN model \cite{hidden} with 3DGS, hereafter referred to as \textbf{HiDDeN + 3DGS}. In this baseline, the HiDDeN model is utilized to watermark the training images, which are then used to optimize the 3DGS. We pre-train the HiDDeN models using the same configuration as that employed for the pre-training of our 2D decoder.
For 3D representation, we select the state-of-the-art published works, \textbf{WateRF} \cite{waterf} for NeRF, and \textbf{GaussianMarker} \cite{NEURIPS2024_39cee562} for 3DGS. The mesh-based method \cite{mesh_watermark1} is excluded due to the significant differences in encoding and rendering compared to 3DGS, and the lack of open-source code.


\subsection{Main Results}
\paragraph{Quantitative comparison.}
We quantitatively compare our framework with all baseline methods on both the source domain (Objaverse$^{GS}_{test}$) and target domain (Blender$^{GS}$, Omni3DObject$^{GS}$) datasets, evaluated on an A100 GPU. The results are presented in \cref{tab:comparison}.
Note that our trained model is not fine-tuned on the target domain datasets, Blender$^{GS}$, Omni3DObject$^{GS}$, underscoring the model's adaptability.
The experimental results demonstrate that our method achieves admirable results on all test datasets, which demonstrates the generalization of our method. In addition, our method achieves better performance than  the NeRF watermarking methods. For the optimization-based 3DGS watermarking method GaussianMarker, although the accuracy of our method is slightly lower than that in the target domain, our method achieves much better visual quality. More importantly, we compute the average watermark embedding time of the test dataset, our method requires significantly less time to embed the messages compared with all the other methods, highlighting the efficiency of our method. 



\begin{figure*}
  \centering
   \includegraphics[width=0.99\linewidth]{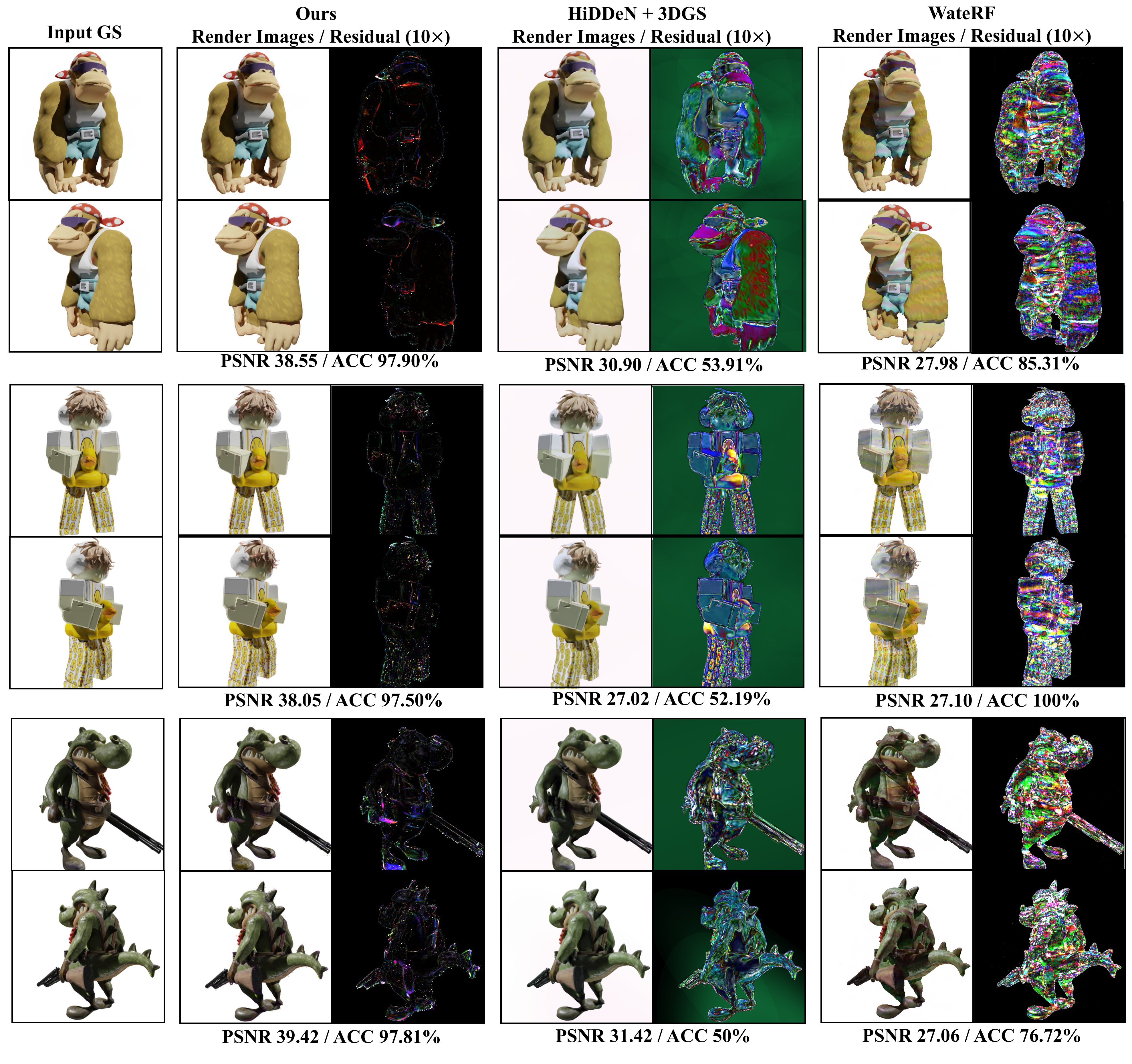}
   \caption{Qualitative comparisons between our method and the baseline. We show the differences (×10) between the images rendered by the input and watermarked 3DGS. Our method achieves better PSNR and bit accuracy than baseline.}
  \label{fig:comparison_baseline}
\end{figure*}

\begin{figure}
  \centering
   \includegraphics[width=0.99\linewidth]{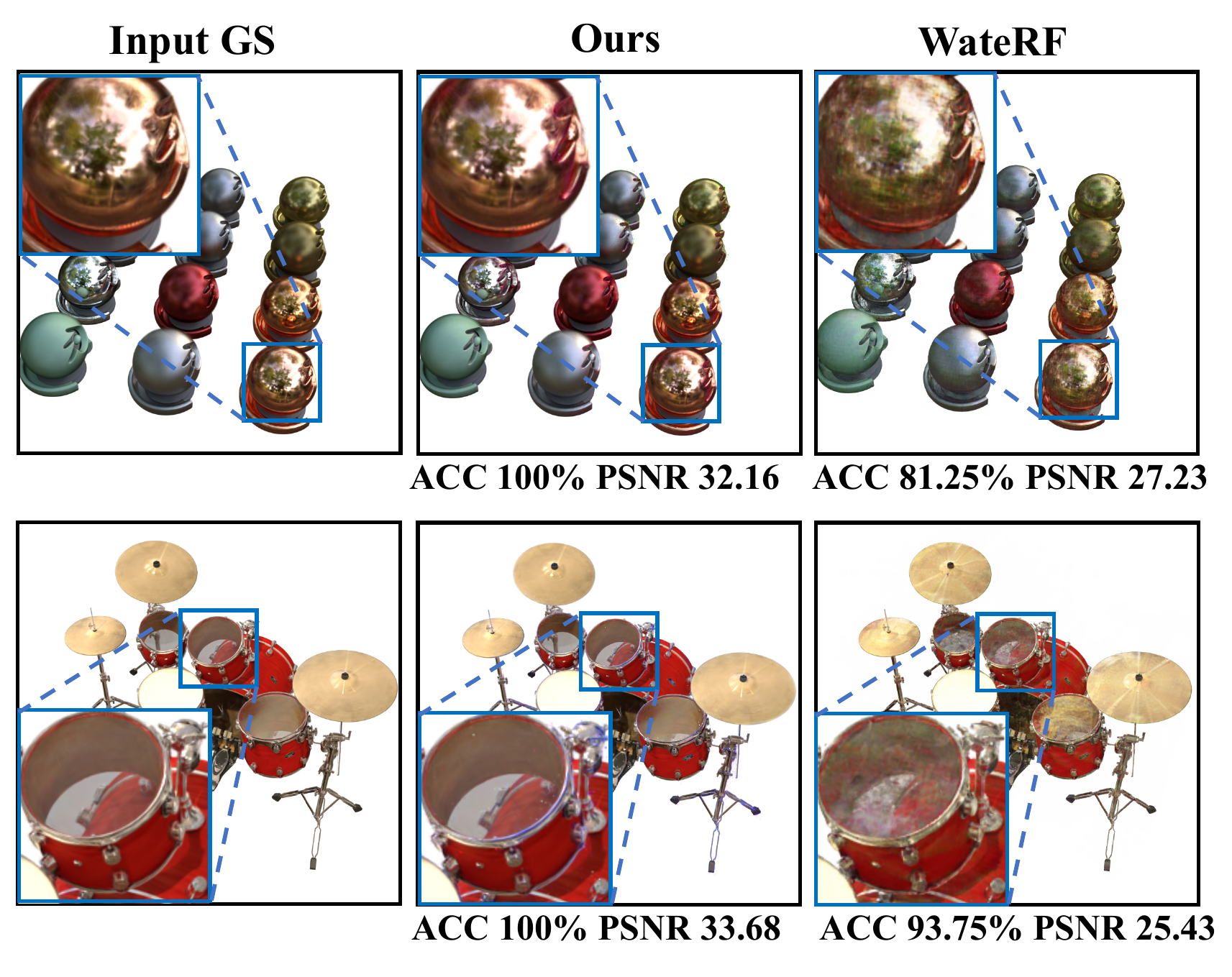}
   \caption{Qualitative comparisons between our method and WateRF \cite{waterf} on the Blender dataset \cite{nerf}. Our method achieves better visual quality than WateRF.}
  \label{fig:comparison_blender}
\end{figure}

\paragraph{Qualitative comparison.}
The qualitative comparison is illustrated in  \cref{fig:comparison_baseline} and \cref{fig:comparison_blender}. 
The discrepancies between the input and watermarked 3DGS in our method are minimal, indicating that watermark messages are embedded with negligible impact on visual fidelity. By contrast, the {HiDDeN + 3DGS} baseline exhibits large, dark Gaussian primitives in the background of rendered images. This effect arises from using HiDDeN’s encoder on the 3DGS training images, which injects invisible noise into the image background. The 3DGS optimization process then attempts to fit these background noises to learn the watermark message, ultimately producing conspicuous Gaussian artifacts.

Meanwhile, WateRF achieves good bit accuracy but introduces noticeable artifacts on the surfaces of watermarked objects, reducing overall visual quality. These results highlight the superior balance our method strikes between embedding robustness and visual realism.

\if false
\begin{table*}[ht!]
\centering
\caption{caption}
\label{tab:comparison}
\begin{tabular}{lccccccccccc}
\toprule
\multirow{2}{*}{Method} 
& \multicolumn{5}{c}{ObjaverseGS$_{test}$} 
& \multicolumn{5}{c}{BlenderGS} 
& \multicolumn{5}{c}{Omni3DObjectGS} \\
\cmidrule(lr){2-6}\cmidrule(lr){7-11}\cmidrule(lr){12-16}
& E-time (s) & Acc (\%) & PSNR$\uparrow$ & SSIM$\uparrow$ & LPIPS$\downarrow$
& E-time (s) & Acc (\%) & PSNR$\uparrow$ & LPIPS$\uparrow$ & SSIM$\downarrow$
& E-time (s) & Acc (\%) & PSNR$\uparrow$ & LPIPS$\uparrow$ & SSIM$\downarrow$ \\
\midrule
HiDDeN+3DGS
& 1 & 1 & 22.64 & 0.21 & 0.75 
& 19.43 & 0.39 & 0.56 
& 22.49 & 0.49 & 0.60 \\
WateRF
& 23.54 & 0.13 & 0.90 
& 19.80 & 0.30 & 0.62 
& 23.27 & 0.48 & 0.69 \\
GaussianMarker
& 23.46 & 0.20 & 0.86 
& 19.70 & 0.26 & 0.60 
& 26.33 & 0.38 & 0.70 \\
\ours
& 22.42 & 0.17 & 0.80 
& 20.54 & 0.24 & 0.64 
& 23.19 & 0.47 & 0.66 \\
\bottomrule
\end{tabular}
\end{table*}
\fi

\begin{table*}[ht!]
\centering
\caption{Quantitative comparison. ``E-time'' is short for average watermark embeddding time tested on an A100 GPU. 
The best values are highlighted in \textbf{bold}, while the second-best accuracy and PSNR are indicated with an \underline{underline}. Note that WateRF and GaussianMarker are iterative optimization-based methods.
}
\label{tab:comparison}
\resizebox{\textwidth}{!}{  
\begin{tabular}{l ccccc ccccc ccccc}
\toprule
\multirow{2}{*}{\textbf{Method}}  
& \multicolumn{5}{c}{\textbf{Objaverse$^{GS}_{test}$ (Source domain)}} 
& \multicolumn{5}{c}{\textbf{Blender$^{GS}$ (Target domain)}} 
& \multicolumn{5}{c}{\textbf{Omni3DObject$^{GS}$ (Target domain)}} \\
\cmidrule(lr){2-6} \cmidrule(lr){7-11} \cmidrule(lr){12-16}
& E-time (s)$\downarrow$ & Acc$\uparrow$ & PSNR$\uparrow$& SSIM$\uparrow$  & LPIPS$\downarrow$ 
& E-time (s)$\downarrow$ & Acc$\uparrow$ & PSNR$\uparrow$& SSIM$\uparrow$  & LPIPS$\downarrow$ 
& E-time (s)$\downarrow$ & Acc$\uparrow$ & PSNR$\uparrow$& SSIM$\uparrow$  & LPIPS$\downarrow$\\
\midrule
Hidden \cite{hidden}+3DGS & 260 & 52.88 & 29.71 & 0.9769 & 0.0459
& 350 & 49.99 & 14.53 & 0.7440 & 0.3671
& 300 & 50.0 & 17.06 & 0.8344 & 0.2889 \\

WateRF \cite{waterf} & 4400 & 83.43 & 33.07 & 0.9626 & 0.0259 
& 4900 & 93.26 & 30.81 & 0.9472 & 0.0577 
& 6000 & 79.54 & 31.16 &  0.9211 & 0.0978 \\

GaussianMarker \cite{NEURIPS2024_39cee562} & 580 & \underline{90.69} & \textbf{35.53} & 0.9760 & 0.0281 
& 340 & \textbf{98.32} & 28.37 & 0.9295 & 0.0516
& 290 & \textbf{94.07} & 31.56 & 0.9413 & 0.0643 \\

\ours~(ours) & \textbf{0.11} & \textbf{95.46} & \underline{34.86}& \textbf{0.9837} & \textbf{0.0249}
& \textbf{0.33} & \underline{93.56} & \textbf{33.40} & \textbf{0.9828} & \textbf{0.0154}
& 0.04& \underline{93.63} & \textbf{37.59} & \textbf{0.9743} & \textbf{0.0229} \\
\bottomrule
\end{tabular}
}
\end{table*}

\subsection{Ablation Study}
\label{sec:ablation}

\paragraph{Impact of modules.}

We perform an ablation study to assess the impact of two main components: the Decoder Fine-Tuning strategy (DF) and the Adaptive Marker Control (AMC) module from \cref{sec:amc}. 
Within AMC, we also examine whether making the perturbation amplitude learnable further enhances performance. The results are summarized in \cref{tab:ablation_module}.
The first row shows that without fine-tuning the 2D decoder, both bit accuracy and visual quality degrade significantly.
Without AMC, the rendering fidelity remains high since the original 3DGS is well-optimized for rendering loss. However, the model fails to improve decoding accuracy due to being trapped in a local minimum of the rendering objective.
Introducing perturbations with fixed amplitude (AMC-F) enhances decoding accuracy while maintaining high rendering fidelity. Further allowing the perturbation amplitude to be learnable (AMC-L) results in additional gains in both accuracy and visual quality. This demonstrates that adaptive control over perturbations is crucial for balancing watermark embedding and rendering quality.

\cref{fig:training_loss} illustrates the evolution of watermark loss, rendering loss, and total loss over iterations for different settings. The without-AMC setting struggles to minimize message loss, as it gets trapped in a local minimum dictated by rendering loss. The AMC-F setting improves message loss decay but fails to optimize rendering loss effectively. In contrast, our AMC-L setting achieves the best balance between watermark accuracy and rendering fidelity, demonstrating its ability to optimize both objectives simultaneously.

The final learned perturbation amplitudes are $\omega_{color}=1.489, \omega_{opacity}=0.2103, \omega_{position}=0.0163, \omega_{scale}=9.9999e-5, \omega_{rotation}=0.0929$.
These values confirm how differentiable rendering in a 3DGS framework influences each parameter. Opacity and color affect the gradient in a nearly linear manner. Position gradients scale with distance from the Gaussian center, normalized by its spread. Scale and rotation are more sensitive, since they depend on squared distances and can dramatically impact Gaussians with small spreads. Overall, this analysis provides insight into how each property contributes to the gradient when 3DGS differentiable rendering is integrated into the learning loop.

\begin{figure}
  \centering
   \includegraphics[width=0.9\linewidth]{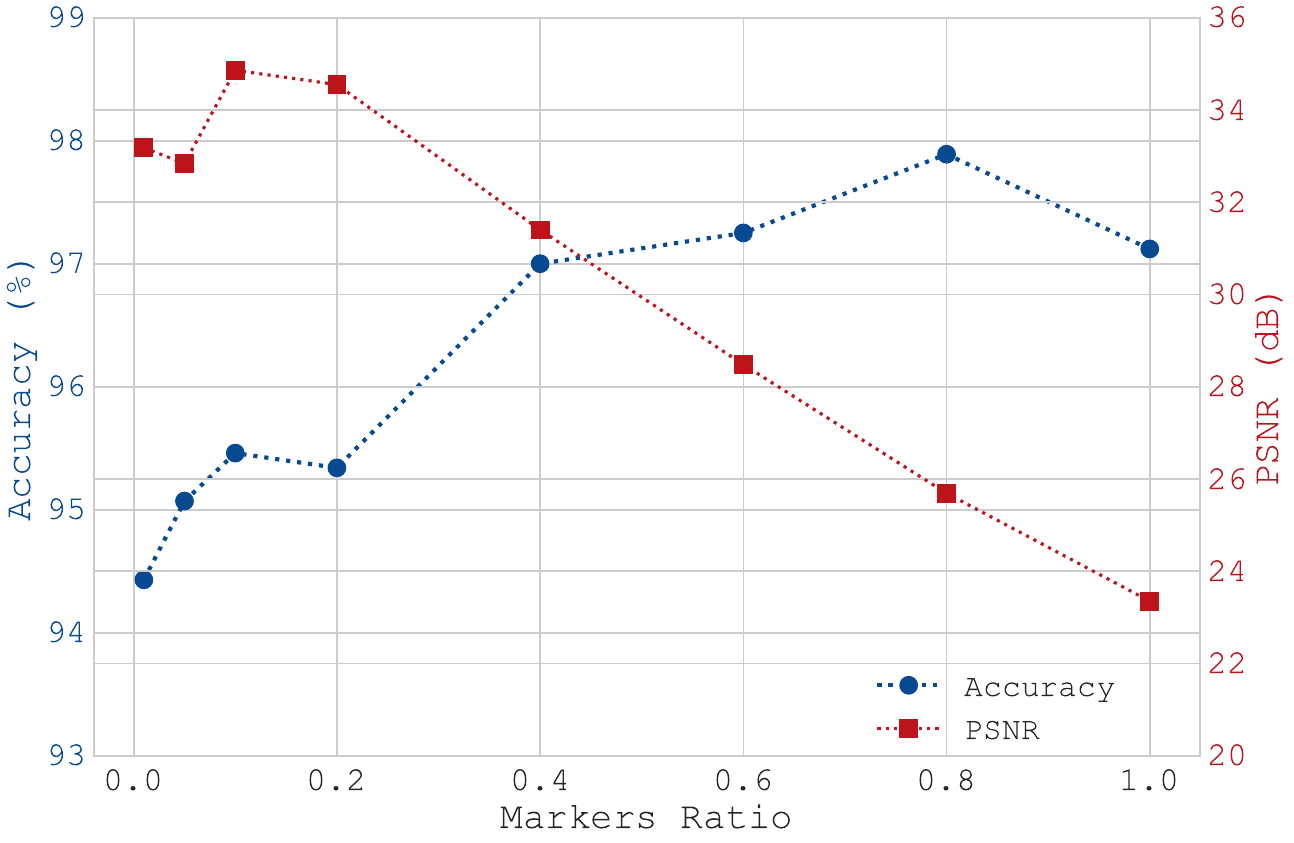}
   \caption{The bit accuracy (\%) and visual qualities of our method under various Markers Ratio.}
  \label{fig:marker_ratio}
\end{figure}

\paragraph{Ablation on Marker Ratio $\gamma$.} 
We investigate how varying the ratio $\gamma$ of selected markers for perturbation affects the model’s performance. A low $\gamma$ risks remaining stuck in local minima, yielding suboptimal accuracy and visual quality. Conversely, a high $\gamma$ excessively disturbs the 3DGS structure, which can degrade rendering quality and increase training difficulty.
As shown in \cref{fig:marker_ratio}, initially increasing $\gamma$ helps the 3DGS escape local minima, improving both bit accuracy and PSNR. However, pushing $\gamma$ too high eventually lowers visual quality despite a further gain in accuracy. The excessive perturbation disrupts the original 3DGS structure, making it difficult for the 3D encoder to maintain high-fidelity rendering. Based on these observations, a moderate perturbation ratio strikes the best balance between watermark accuracy and visual quality. We select $\gamma=0.4$ as our final setting.

\if false
\begin{table}
    \begin{center}
    \setlength{\tabcolsep}{2.5pt}
    \resizebox{\linewidth}{!}{
    \begin{tabular}{ccccc}
    \toprule[1pt]
     Markers Ratio & Acc $\uparrow$ & PSNR $\uparrow$ & SSIM $\uparrow$ & LPIPS $\downarrow$ \\
    \hline
    0.0 & 90.14 & 31.61 & 0.9760 & 0.0366 \\
    \hline
    0.01 & 94.43 & 33.18 & 0.9848 & 0.0237 \\
    \hline
    0.05 & 95.07 & 32.84 & 0.9815 & 0.0302 \\
    \hline
    0.1 & 95.46 & 34.86 & 0.9837 & 0.0249 \\
    \hline
    0.2 & 95.34 & 34.55 & 0.9768 & 0.0327 \\
    \hline
    0.4 & 97.00 & 31.40 & 0.9630 & 0.0499 \\
    \hline
    0.6 & 97.25 & 28.48 & 0.9402 & 0.0773 \\
    \hline
    0.8 & 97.89 & 25.69 & 0.9147 & 0.1114 \\
    \hline
    1.0 & 97.12 & 23.35 & 0.8859 & 0.1465 \\
    \hline
    \toprule[1pt]
    \end{tabular}
    }
    \end{center}
    \caption{The bit accuracy ($\%$) and visual qualities of our method under various Markers Ratio.}
\label{tab.marker_ratio}
\end{table}
\fi

\begin{figure*}[ht]
  \centering
  \begin{minipage}{0.73\textwidth}  
    \centering
    \includegraphics[width=0.32\linewidth]{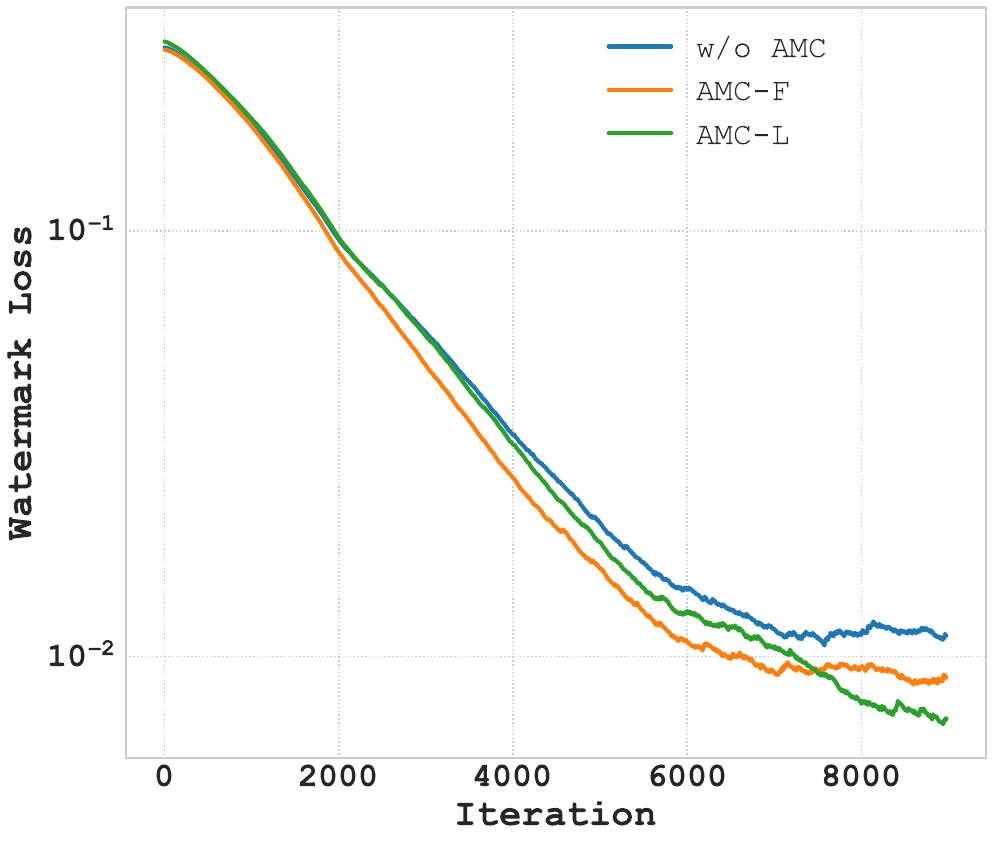} \hfill%
    \includegraphics[width=0.32\linewidth]{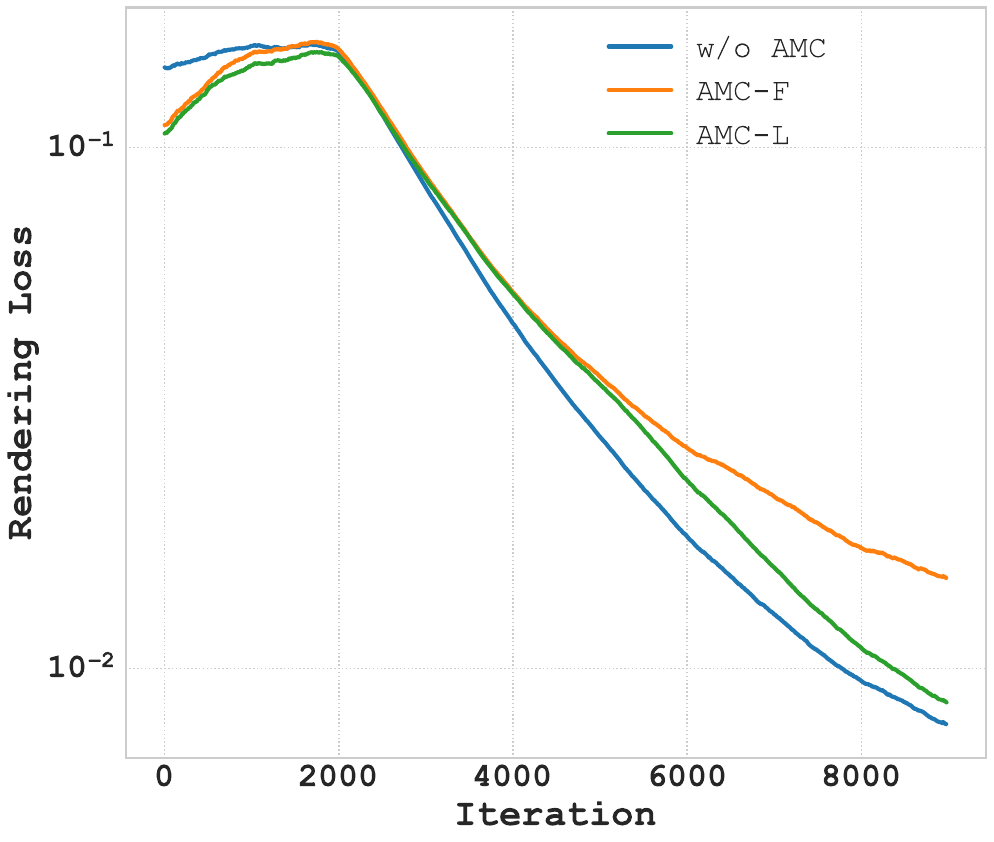} \hfill%
    \includegraphics[width=0.32\linewidth]{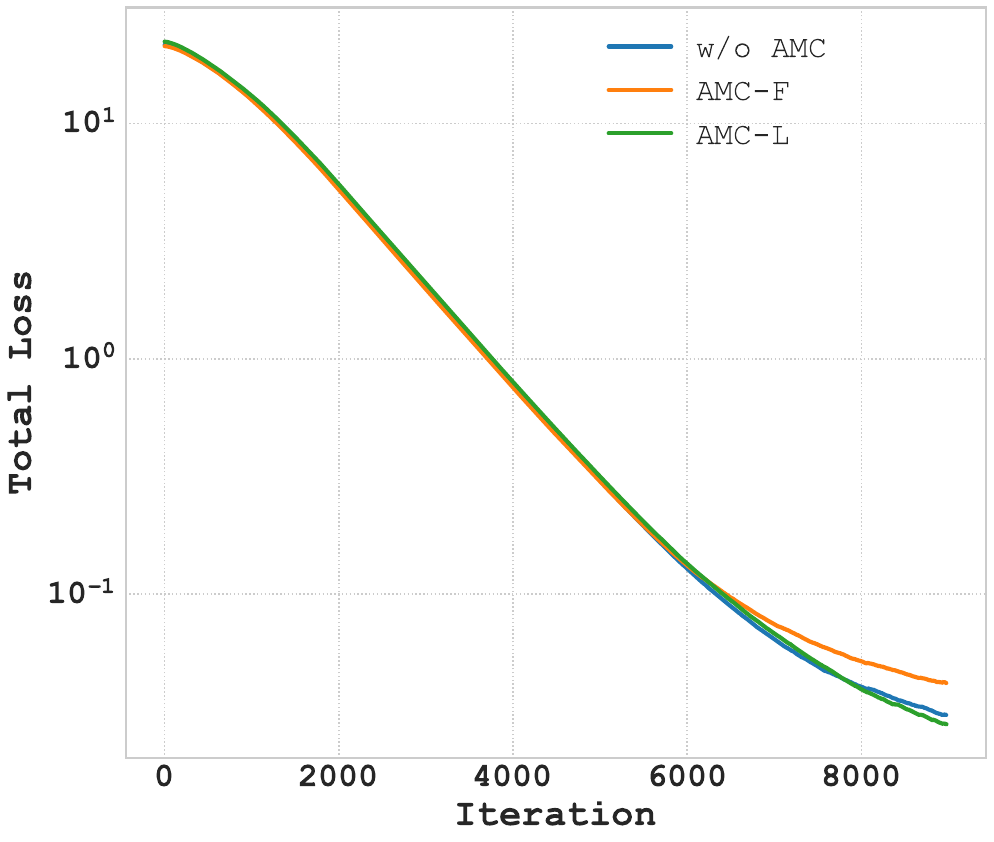}
    \caption{
      The evolution of watermark, rendering, and total loss during training under different settings. ``AMC-F'' and ``AMC-L'' denote different settings. Ours ``AMC-L'' best balances the losses.
    }
    \label{fig:training_loss}
  \end{minipage}%
  \hfill%
  \begin{minipage}{0.243\textwidth}  
    \centering
    \includegraphics[width=\linewidth]{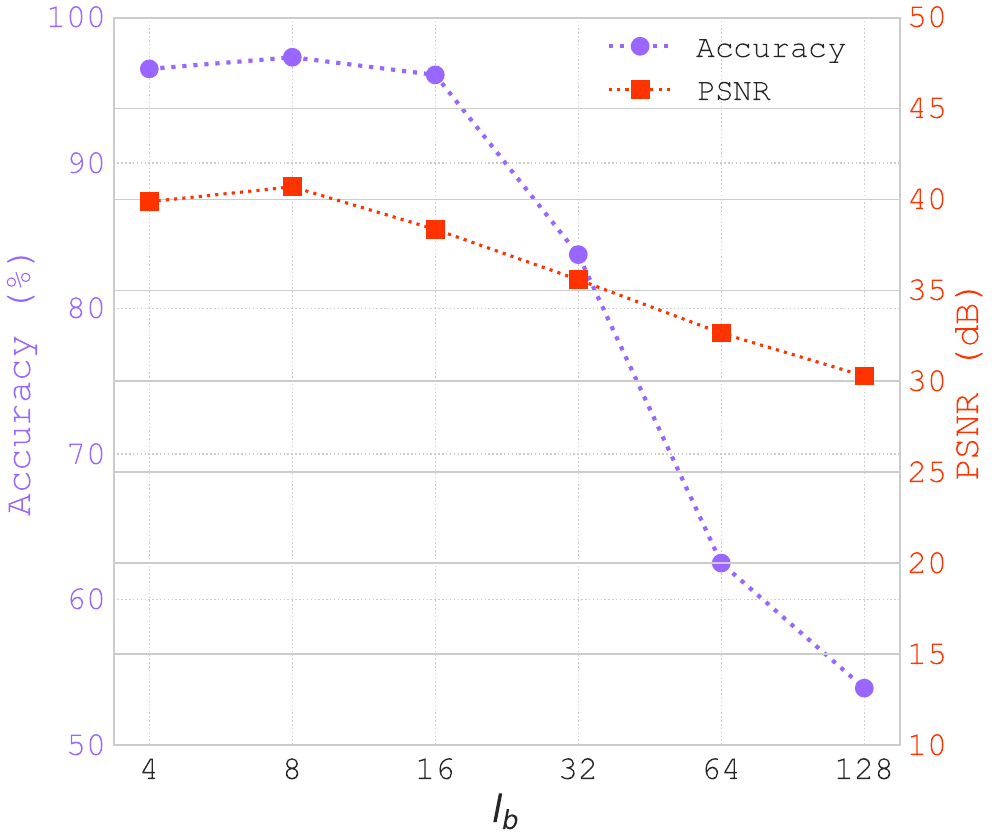}
    \caption{The model performance under various message lengths $l_b$.}
    \label{fig:ablation_studies}
  \end{minipage}
  \vspace{-8pt}
\end{figure*}

\begin{table}[!t]
    \centering
    \setlength{\tabcolsep}{2.5pt} 
    \renewcommand{\arraystretch}{1.2} 
    \newcolumntype{C}{>{\centering\arraybackslash}p{0.75cm}} 
    \newcolumntype{M}{>{\centering\arraybackslash}p{1.3cm}} 
    \newcolumntype{L}{>{\centering\arraybackslash}X}        
    
    \begin{tabularx}{\columnwidth}{CC|M L L L}
    \toprule[1pt]
    FD & AMC & Acc $\uparrow$ & PSNR $\uparrow$ & SSIM $\uparrow$ & LPIPS $\downarrow$ \\
    \midrule
    \ding{55} &               & 52.86  & 21.93  & 0.9342 & 0.0914 \\
              & \ding{55}             & 90.14  & \textbf{31.61} & \textbf{0.9760} & \textbf{0.0366} \\
              & F     & 94.96  & 31.24  & 0.9667 & 0.0568 \\
              & L             & \textbf{96.68} & 31.55 & 0.9643 & 0.0484 \\
    \bottomrule[1pt]
    \end{tabularx}
    \vspace{-6pt}
    \caption{Impact on modules. 
``FD" and ``AMC" represent ``fine-tuning decoder" and ``adaptive marker control," respectively. In the AMC column, ``F" indicates a fixed perturbation amplitude, while ``L" denotes a learnable perturbation amplitude.
    }
    \label{tab:ablation_module}
\end{table}


\begin{figure}
  \centering
   \includegraphics[width=0.99\linewidth]{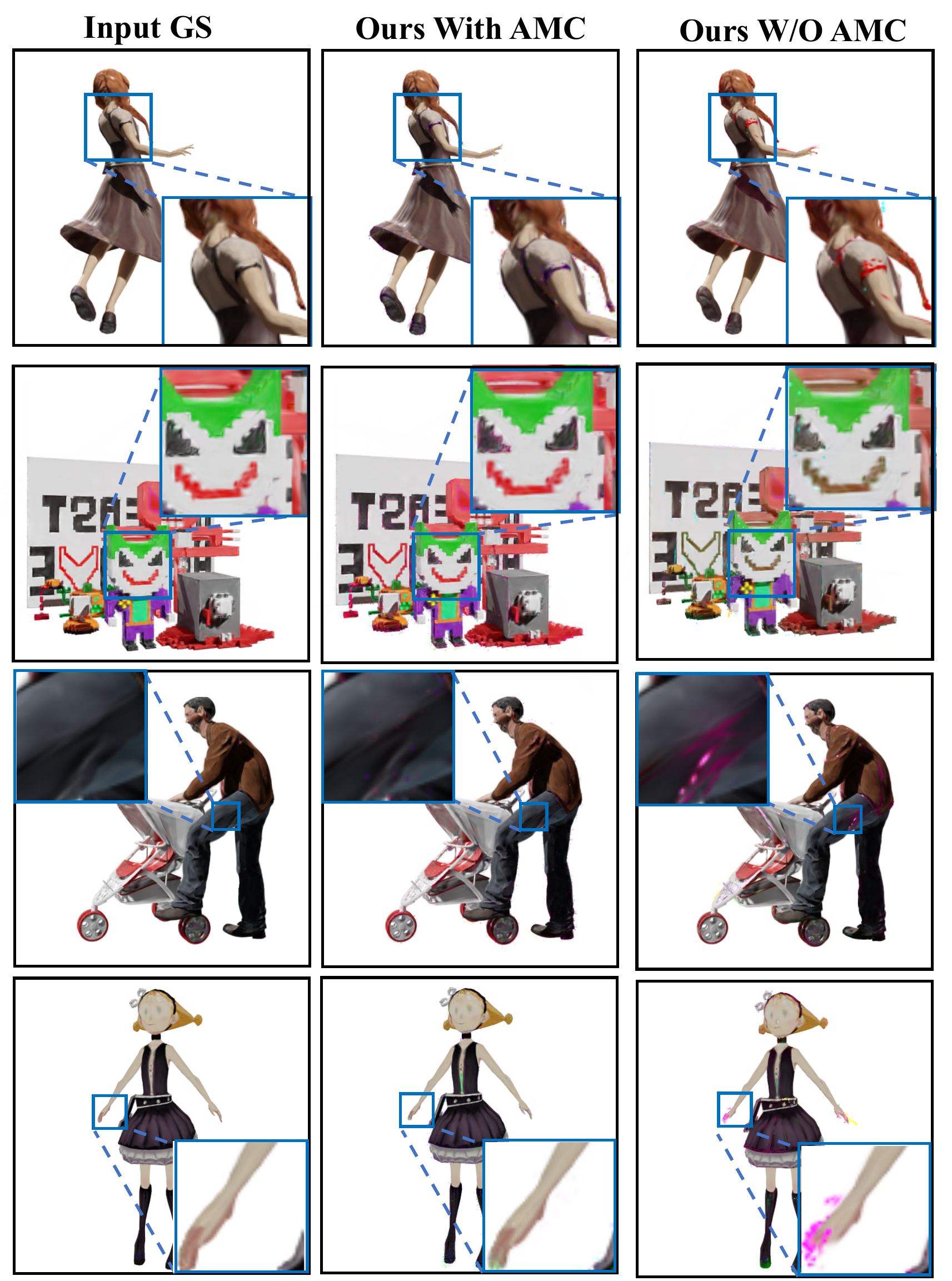}
   \caption{Qualitative comparisons between our method with and without Adaptive Marker Control (AMC). Without AMC, artifacts appear on the rendered images. }
  \label{fig:ablation_noise}
\end{figure}

\if false
\begin{table}
    \begin{center}
    \setlength{\tabcolsep}{2.5pt}
    \begin{tabular}{ccccc}
    \toprule[1pt]
    $l_b$ & Acc ($\%$) $\uparrow$ & PSNR $\uparrow$ & SSIM $\uparrow$ & LPIPS $\downarrow$ \\
    \hline
   4 & 96.48 & 39.88 & 0.9938 & 0.0103 \\
   8 & 97.28 & 40.70 & 0.9940 & 0.0087 \\
   16 & 96.07 & 38.36 & 0.9898 & 0.0149 \\
   32 & 83.71 & 35.59 & 0.9767 & 0.0402 \\
    \hline
    \toprule[1pt]
    \end{tabular}
    \end{center}
    \vspace{-10pt}
    \caption{The model performance under various message length $l_b$
    \ca{separate, only length}
    }
\label{tab.ablation_studies}
\vspace{-6pt}
\end{table}
\fi

\if false
\begin{table}
    \centering
    \setlength{\tabcolsep}{4pt} 
    \renewcommand{\arraystretch}{1.2} 
    \newcolumntype{Y}{>{\centering\arraybackslash}X} 
    \newcolumntype{Z}{>{\centering\arraybackslash}p{2.2cm}} 
    \begin{tabularx}{\columnwidth}{Y Z Y Y Y}  
    \toprule[1pt]
    $l_b$ & Acc ($\%$) $\uparrow$ & PSNR $\uparrow$ & SSIM $\uparrow$ & LPIPS $\downarrow$ \\
    \midrule
    4  & 96.48  & 39.88  & 0.9938  & 0.0103 \\
    8  & \textbf{97.28}  & \textbf{40.70}  & \textbf{0.9940}  & \textbf{0.0087} \\
    16 & 96.07  & 38.36  & 0.9898  & 0.0149 \\
    32 & 83.71  & 35.59  & 0.9767  & 0.0402 \\
    64 & 62.50 & 22.65 & 0.8658 & 0.1662 \\
    128 & 53.90 & 30.29 & 0.9652 & 0.0668 \\
    \bottomrule[1pt]
    \end{tabularx}
    \vspace{-6pt}
    \caption{The model performance under various message lengths $l_b$ \textit{(separate, only length)}.}
    \label{tab.ablation_studies}
\end{table}
\fi


\if false
\begin{figure*}[ht]
  \centering
  \begin{subfigure}[t]{0.24\textwidth}  
    \centering
    \includegraphics[width=\linewidth]{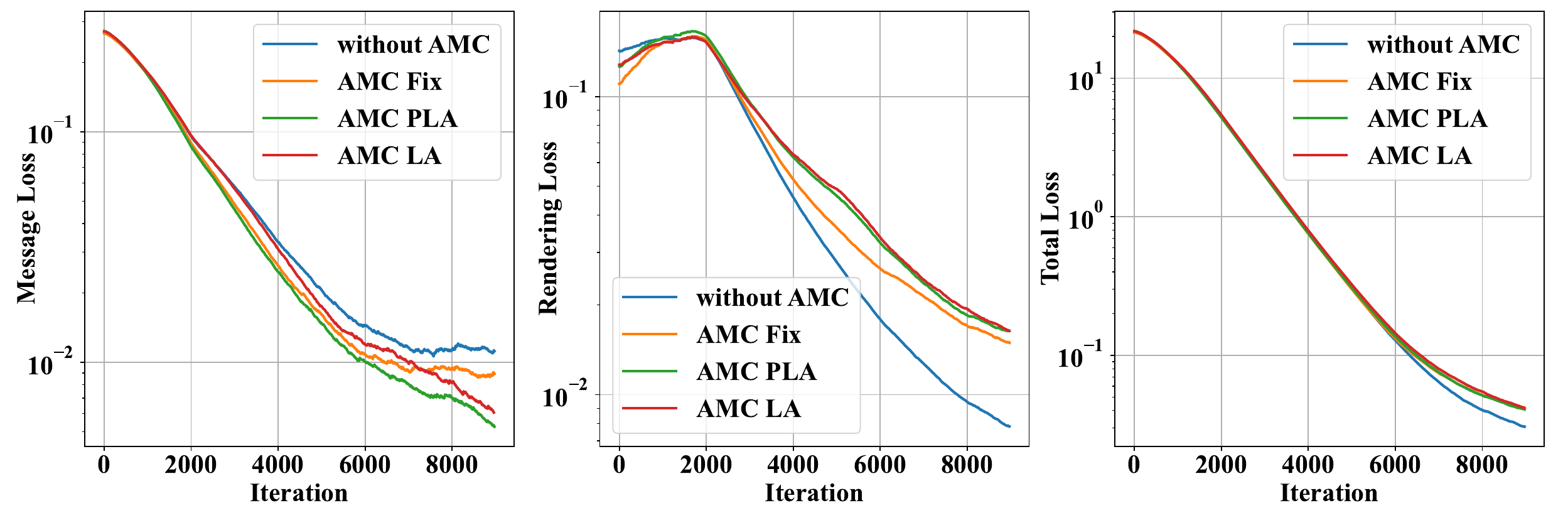}
    \caption{Training loss of the models under various settings. ``Fix'', ``PLA'', and ``LA'' denote three types of learnable perturbation amplitude settings: fix amplitude, partial learnable amplitude, learnable amplitude.}
    \label{fig:training_loss}
  \end{subfigure}%
  \hfill
  \begin{subfigure}[t]{0.335\textwidth}  
    \centering
    \includegraphics[width=\linewidth]{Figs/lb_vs_acc_psnr.pdf}
    \caption{The model performance under various message lengths $l_b$ \textit{(separate, only length)}.}
    \label{fig:bit_length}
  \end{subfigure}
  \vspace{-8pt}
\end{figure*}
\fi

\paragraph{Message length.}

We conduct a series of experiments to analyze the impact of message length $l_b$ to explore the embedding capacity of our model.
As shown in \cref{fig:ablation_studies}, both bit accuracy and visual quality decline as $l_b$ increases beyond 32. 
For shorter message ($l_b < 32$), our model maintains high accuracy and good visual quality. This indicates that while longer messages introduce more embedding complexity, moderate message lengths strike a balance between watermark robustness and rendering fidelity.

\subsection{Distortion impact.}

We evaluate the robustness of the proposed \ours~method against various distortions, as presented in \cref{tab.distortion}. 
The results demonstrate that our method maintains high bit accuracy across a wide range of both 2D and 3D distortions. While minor variations exist among different distortion types, the overall performance remains stable, indicating strong resilience. Notably, the model achieves 96.03\% accuracy under dropout distortion (2D) and 96.31\% under crop distortion (3D), further confirming its robustness in challenging conditions.



\begin{table}[htbp]
    \begin{center}
    \setlength{\tabcolsep}{1.5pt}
    \small
    \resizebox{0.99\linewidth}{!}{

    \begin{tabular}{c|ccccc|cc}
    \toprule[1pt]
       \multirow{2}{*}{None}  & \multicolumn{5}{|c|}{\textbf{2D Distortion}} & \multicolumn{2}{c}{\textbf{3D Distortion}} \\
      & \makecell{Crop \\[-3pt] \scriptsize $cr=0.4$} & \makecell{Dropout \\[-3pt] \scriptsize $p=0.1$} & \makecell{JPEG \\[-3pt] \scriptsize $Q=50$}  & \makecell{Rotation \\[-3pt] \scriptsize $r=\pm30^{\circ}$} & \makecell{Noise \\[-3pt] \footnotesize $\sigma=0.06$}  & \makecell{Dropout \\[-3pt] \scriptsize $p=0.2$} & \makecell{Cropout \\[-3pt] \scriptsize $cr=0.4$} \\
    \hline
    97.67 & 95.02 & 96.03 & 94.60 & 96.77 & 96.54 & 95.91 & 96.31 \\
    \textit{Delta} & \textit{-2.65} & \textit{-1.64} & \textit{-3.07} & \textit{-0.90} & \textit{-1.13} & \textit{-1.76} & \textit{-1.36} \\
    
    \toprule[1pt]
    \end{tabular}         
    }
    \end{center}
    \vspace{-20pt}
    \caption{Bit accuracy under various distortions.}
\label{tab.distortion}
\vspace{-15pt}
\end{table}

\section{Conclusion}
We present a robust, generalizable watermarking framework for 3D Gaussian Splatting (3DGS). It embeds watermark information into the 3DGS in a single forward pass and extracts the watermark from rendered images. Our framework comprises a 3D encoder to insert watermarks into the 3DGS, distortion layers to enhance robustness, and a 2D decoder to recover messages from the rendered images. Although the input 3DGS is already optimized for rendering, it often gets trapped in local minima that hinder watermark accuracy. To address this, we introduce an adaptive marker control mechanism that helps the model escape these local minima, improving training stability and convergence. This enables effective and efficient watermark embedding for 3DGS in a single forward pass.
\vspace{-10pt}
\paragraph{Limitation.}
A major limitation of our approach is that each Gaussian primitive integrates message bits, which can lead to redundancy and increased computational costs, thereby limiting the embedding capacity. Future work will focus on developing a more efficient embedding network that enhances embedding capacity while reducing redundancy.



{
    \small
    \bibliographystyle{ieeenat_fullname}
    \bibliography{main}

\begin{thebibliography}{36}
\providecommand{\natexlab}[1]{#1}
\providecommand{\url}[1]{\texttt{#1}}
\expandafter\ifx\csname urlstyle\endcsname\relax
  \providecommand{\doi}[1]{doi: #1}\else
  \providecommand{\doi}{doi: \begingroup \urlstyle{rm}\Url}\fi

\bibitem[Chu and Harada(2024)]{3dgs_avatars2}
Xuangeng Chu and Tatsuya Harada.
\newblock Generalizable and animatable gaussian head avatar.
\newblock \emph{arXiv preprint arXiv:2410.07971}, 2024.

\bibitem[Deitke et~al.(2023)Deitke, Schwenk, Salvador, Weihs, Michel, VanderBilt, Schmidt, Ehsani, Kembhavi, and Farhadi]{objaverse}
Matt Deitke, Dustin Schwenk, Jordi Salvador, Luca Weihs, Oscar Michel, Eli VanderBilt, Ludwig Schmidt, Kiana Ehsani, Aniruddha Kembhavi, and Ali Farhadi.
\newblock Objaverse: A universe of annotated 3d objects.
\newblock In \emph{Proceedings of the IEEE/CVF Conference on Computer Vision and Pattern Recognition}, pages 13142--13153, 2023.

\bibitem[Fernandez et~al.(2023)Fernandez, Couairon, J{\'e}gou, Douze, and Furon]{image_watermark3}
Pierre Fernandez, Guillaume Couairon, Herv{\'e} J{\'e}gou, Matthijs Douze, and Teddy Furon.
\newblock The stable signature: Rooting watermarks in latent diffusion models.
\newblock In \emph{Proceedings of the IEEE/CVF International Conference on Computer Vision}, pages 22466--22477, 2023.

\bibitem[Ferreira and Lima(2020)]{point_watermark}
Felipe~ABS Ferreira and Juliano~B Lima.
\newblock A robust 3d point cloud watermarking method based on the graph fourier transform.
\newblock \emph{Multimedia Tools and Applications}, 79\penalty0 (3):\penalty0 1921--1950, 2020.

\bibitem[Fu et~al.(2024)Fu, Wang, Liu, Kulkarni, Kautz, and Efros]{3dgs_scene2}
Yang Fu, Xiaolong Wang, Sifei Liu, Amey Kulkarni, Jan Kautz, and Alexei~A. Efros.
\newblock Colmap-free 3d gaussian splatting.
\newblock In \emph{2024 IEEE/CVF Conference on Computer Vision and Pattern Recognition (CVPR)}, pages 20796--20805, 2024.

\bibitem[He et~al.(2023)He, Wang, Zhang, Abdullahi, and Yang]{video_watermark2}
Mingze He, Hongxia Wang, Fei Zhang, Sani~M. Abdullahi, and Ling Yang.
\newblock Robust blind video watermarking against geometric deformations and online video sharing platform processing.
\newblock \emph{IEEE Transactions on Dependable and Secure Computing}, 20\penalty0 (6):\penalty0 4702--4718, 2023.

\bibitem[Huang et~al.(2024)Huang, Li, Cheung, Cheung, See, and Wan]{NEURIPS2024_39cee562}
Xiufeng Huang, Ruiqi Li, Yiu-ming Cheung, Ka~Chun Cheung, Simon See, and Renjie Wan.
\newblock Gaussianmarker: Uncertainty-aware copyright protection of 3d gaussian splatting.
\newblock In \emph{Advances in Neural Information Processing Systems}, pages 33037--33060. Curran Associates, Inc., 2024.

\bibitem[Jang et~al.(2024{\natexlab{a}})Jang, Lee, Jang, Kim, Yang, and Kim]{waterf}
Youngdong Jang, Dong~In Lee, MinHyuk Jang, Jong~Wook Kim, Feng Yang, and Sangpil Kim.
\newblock Waterf: Robust watermarks in radiance fields for protection of copyrights.
\newblock In \emph{Proceedings of the IEEE/CVF Conference on Computer Vision and Pattern Recognition}, pages 12087--12097, 2024{\natexlab{a}}.

\bibitem[Jang et~al.(2024{\natexlab{b}})Jang, Park, Yang, Ko, Choo, and Kim]{jang20243d}
Youngdong Jang, Hyunje Park, Feng Yang, Heeju Ko, Euijin Choo, and Sangpil Kim.
\newblock 3d-gsw: 3d gaussian splatting watermark for protecting copyrights in radiance fields.
\newblock \emph{arXiv preprint arXiv:2409.13222}, 2024{\natexlab{b}}.

\bibitem[Kerbl et~al.(2023)Kerbl, Kopanas, Leimk{\"u}hler, and Drettakis]{3dgs}
Bernhard Kerbl, Georgios Kopanas, Thomas Leimk{\"u}hler, and George Drettakis.
\newblock 3d gaussian splatting for real-time radiance field rendering.
\newblock \emph{ACM Trans. Graph.}, 42\penalty0 (4):\penalty0 139--1, 2023.

\bibitem[Kocabas et~al.(2024)Kocabas, Chang, Gabriel, Tuzel, and Ranjan]{3dgs_avatars5}
Muhammed Kocabas, Jen-Hao~Rick Chang, James Gabriel, Oncel Tuzel, and Anurag Ranjan.
\newblock Hugs: Human gaussian splats.
\newblock In \emph{Proceedings of the IEEE/CVF conference on computer vision and pattern recognition}, pages 505--515, 2024.

\bibitem[Li et~al.(2023)Li, Feng, Fan, Pan, and Wang]{nerf_watermark1}
Chenxin Li, Brandon~Y Feng, Zhiwen Fan, Panwang Pan, and Zhangyang Wang.
\newblock Steganerf: Embedding invisible information within neural radiance fields.
\newblock In \emph{Proceedings of the IEEE/CVF International Conference on Computer Vision}, pages 441--453, 2023.

\bibitem[Li et~al.(2024)Li, Zheng, Wang, and Liu]{3dgs_avatars3}
Zhe Li, Zerong Zheng, Lizhen Wang, and Yebin Liu.
\newblock Animatable gaussians: Learning pose-dependent gaussian maps for high-fidelity human avatar modeling.
\newblock In \emph{Proceedings of the IEEE/CVF Conference on Computer Vision and Pattern Recognition}, pages 19711--19722, 2024.

\bibitem[Lin et~al.(2024)Lin, Li, Tang, Liu, Liu, Liu, Lu, Wu, Xu, Yan, et~al.]{3dgs_scene3}
Jiaqi Lin, Zhihao Li, Xiao Tang, Jianzhuang Liu, Shiyong Liu, Jiayue Liu, Yangdi Lu, Xiaofei Wu, Songcen Xu, Youliang Yan, et~al.
\newblock Vastgaussian: Vast 3d gaussians for large scene reconstruction.
\newblock In \emph{Proceedings of the IEEE/CVF Conference on Computer Vision and Pattern Recognition}, pages 5166--5175, 2024.

\bibitem[Lu et~al.(2021)Lu, Wang, Zhong, and Rosin]{image_watermark1}
Shao-Ping Lu, Rong Wang, Tao Zhong, and Paul~L Rosin.
\newblock Large-capacity image steganography based on invertible neural networks.
\newblock In \emph{Proceedings of the IEEE/CVF conference on computer vision and pattern recognition}, pages 10816--10825, 2021.

\bibitem[Luo et~al.(2023{\natexlab{a}})Luo, Li, Chang, Liu, Milanfar, and Yang]{video_watermark3}
Xiyang Luo, Yinxiao Li, Huiwen Chang, Ce Liu, Peyman Milanfar, and Feng Yang.
\newblock Dvmark: A deep multiscale framework for video watermarking.
\newblock \emph{IEEE Transactions on Image Processing}, pages 1--1, 2023{\natexlab{a}}.

\bibitem[Luo et~al.(2023{\natexlab{b}})Luo, Guo, Cheung, See, and Wan]{copyrnerf}
Ziyuan Luo, Qing Guo, Ka~Chun Cheung, Simon See, and Renjie Wan.
\newblock Copyrnerf: Protecting the copyright of neural radiance fields.
\newblock In \emph{Proceedings of the IEEE/CVF International Conference on Computer Vision}, pages 22401--22411, 2023{\natexlab{b}}.

\bibitem[Mildenhall et~al.(2021)Mildenhall, Srinivasan, Tancik, Barron, Ramamoorthi, and Ng]{nerf}
Ben Mildenhall, Pratul~P Srinivasan, Matthew Tancik, Jonathan~T Barron, Ravi Ramamoorthi, and Ren Ng.
\newblock Nerf: Representing scenes as neural radiance fields for view synthesis.
\newblock \emph{Communications of the ACM}, 65\penalty0 (1):\penalty0 99--106, 2021.

\bibitem[Moreau et~al.(2024)Moreau, Song, Dhamo, Shaw, Zhou, and P{\'e}rez-Pellitero]{3dgs_avatars4}
Arthur Moreau, Jifei Song, Helisa Dhamo, Richard Shaw, Yiren Zhou, and Eduardo P{\'e}rez-Pellitero.
\newblock Human gaussian splatting: Real-time rendering of animatable avatars.
\newblock In \emph{Proceedings of the IEEE/CVF Conference on Computer Vision and Pattern Recognition}, pages 788--798, 2024.

\bibitem[Mou et~al.(2023)Mou, Xu, Song, Zhao, Ghanem, and Zhang]{video_watermark1}
Chong Mou, Youmin Xu, Jiechong Song, Chen Zhao, Bernard Ghanem, and Jian Zhang.
\newblock Large-capacity and flexible video steganography via invertible neural network.
\newblock In \emph{Proceedings of the IEEE/CVF Conference on Computer Vision and Pattern Recognition}, pages 22606--22615, 2023.

\bibitem[Qi et~al.(2017)Qi, Yi, Su, and Guibas]{pointnet++}
Charles~Ruizhongtai Qi, Li Yi, Hao Su, and Leonidas~J Guibas.
\newblock Pointnet++: Deep hierarchical feature learning on point sets in a metric space.
\newblock \emph{Advances in neural information processing systems}, 30, 2017.

\bibitem[Shao et~al.(2024)Shao, Wang, Li, Wang, Lin, Zhang, Fan, and Wang]{3dgs_avatars1}
Zhijing Shao, Zhaolong Wang, Zhuang Li, Duotun Wang, Xiangru Lin, Yu Zhang, Mingming Fan, and Zeyu Wang.
\newblock Splattingavatar: Realistic real-time human avatars with mesh-embedded gaussian splatting.
\newblock In \emph{Proceedings of the IEEE/CVF Conference on Computer Vision and Pattern Recognition}, pages 1606--1616, 2024.

\bibitem[Tancik et~al.(2020)Tancik, Mildenhall, and Ng]{image_watermark4}
Matthew Tancik, Ben Mildenhall, and Ren Ng.
\newblock Stegastamp: Invisible hyperlinks in physical photographs.
\newblock In \emph{Proceedings of the IEEE/CVF conference on computer vision and pattern recognition}, pages 2117--2126, 2020.

\bibitem[Tang et~al.(2025)Tang, Chen, Chen, Wang, Zeng, and Liu]{LGM}
Jiaxiang Tang, Zhaoxi Chen, Xiaokang Chen, Tengfei Wang, Gang Zeng, and Ziwei Liu.
\newblock Lgm: Large multi-view gaussian model for high-resolution 3d content creation.
\newblock In \emph{European Conference on Computer Vision}, pages 1--18. Springer, 2025.

\bibitem[Wang et~al.(2022)Wang, Zhou, Fang, Zhang, and Yu]{mesh_watermark2}
Feng Wang, Hang Zhou, Han Fang, Weiming Zhang, and Nenghai Yu.
\newblock Deep 3d mesh watermarking with self-adaptive robustness.
\newblock \emph{Cybersecurity}, 5\penalty0 (1):\penalty0 24, 2022.

\bibitem[Wu et~al.(2023)Wu, Zhang, Fu, Wang, Ren, Pan, Wu, Yang, Wang, Qian, et~al.]{wu2023omniobject3d}
Tong Wu, Jiarui Zhang, Xiao Fu, Yuxin Wang, Jiawei Ren, Liang Pan, Wayne Wu, Lei Yang, Jiaqi Wang, Chen Qian, et~al.
\newblock Omniobject3d: Large-vocabulary 3d object dataset for realistic perception, reconstruction and generation.
\newblock In \emph{Proceedings of the IEEE/CVF Conference on Computer Vision and Pattern Recognition}, pages 803--814, 2023.

\bibitem[Xu et~al.(2022)Xu, Mou, Hu, Xie, and Zhang]{image_watermark5}
Youmin Xu, Chong Mou, Yujie Hu, Jingfen Xie, and Jian Zhang.
\newblock Robust invertible image steganography.
\newblock In \emph{Proceedings of the IEEE/CVF conference on computer vision and pattern recognition}, pages 7875--7884, 2022.

\bibitem[Xu et~al.(2024)Xu, Shi, Yifan, Chen, Yang, Peng, Shen, and Wetzstein]{GRM}
Yinghao Xu, Zifan Shi, Wang Yifan, Hansheng Chen, Ceyuan Yang, Sida Peng, Yujun Shen, and Gordon Wetzstein.
\newblock Grm: Large gaussian reconstruction model for efficient 3d reconstruction and generation.
\newblock \emph{arXiv preprint arXiv:2403.14621}, 2024.

\bibitem[Yoo et~al.(2022)Yoo, Chang, Luo, Stava, Liu, Milanfar, and Yang]{mesh_watermark1}
Innfarn Yoo, Huiwen Chang, Xiyang Luo, Ondrej Stava, Ce Liu, Peyman Milanfar, and Feng Yang.
\newblock Deep 3d-to-2d watermarking: Embedding messages in 3d meshes and extracting them from 2d renderings.
\newblock In \emph{Proceedings of the IEEE/CVF Conference on Computer Vision and Pattern Recognition}, pages 10031--10040, 2022.

\bibitem[Zhang et~al.(2018)Zhang, Isola, Efros, Shechtman, and Wang]{lpips}
Richard Zhang, Phillip Isola, Alexei~A Efros, Eli Shechtman, and Oliver Wang.
\newblock The unreasonable effectiveness of deep features as a perceptual metric.
\newblock In \emph{Proceedings of the IEEE conference on computer vision and pattern recognition}, pages 586--595, 2018.

\bibitem[Zhang et~al.(2024{\natexlab{a}})Zhang, Li, Yu, Xu, Li, and Zhang]{image_watermark2}
Xuanyu Zhang, Runyi Li, Jiwen Yu, Youmin Xu, Weiqi Li, and Jian Zhang.
\newblock Editguard: Versatile image watermarking for tamper localization and copyright protection.
\newblock In \emph{Proceedings of the IEEE/CVF Conference on Computer Vision and Pattern Recognition}, pages 11964--11974, 2024{\natexlab{a}}.

\bibitem[Zhang et~al.(2024{\natexlab{b}})Zhang, Meng, Li, Xu, Zhang, and Zhang]{gs_hider}
Xuanyu Zhang, Jiarui Meng, Runyi Li, Zhipei Xu, Yongbing Zhang, and Jian Zhang.
\newblock Gs-hider: Hiding messages into 3d gaussian splatting.
\newblock \emph{arXiv preprint arXiv:2405.15118}, 2024{\natexlab{b}}.

\bibitem[Zhou et~al.(2024)Zhou, Lin, Shan, Wang, Sun, and Yang]{3dgs_scene1}
Xiaoyu Zhou, Zhiwei Lin, Xiaojun Shan, Yongtao Wang, Deqing Sun, and Ming-Hsuan Yang.
\newblock Drivinggaussian: Composite gaussian splatting for surrounding dynamic autonomous driving scenes.
\newblock In \emph{Proceedings of the IEEE/CVF Conference on Computer Vision and Pattern Recognition}, pages 21634--21643, 2024.

\bibitem[Zhu(2018)]{hidden}
J Zhu.
\newblock Hidden: hiding data with deep networks.
\newblock \emph{arXiv preprint arXiv:1807.09937}, 2018.

\bibitem[Zhu et~al.(2024)Zhu, Ye, Luo, and Wei]{mesh_watermark3}
Xingyu Zhu, Guanhui Ye, Xiapu Luo, and Xuetao Wei.
\newblock Rethinking mesh watermark: Towards highly robust and adaptable deep 3d mesh watermarking.
\newblock In \emph{Proceedings of the AAAI Conference on Artificial Intelligence}, pages 7784--7792, 2024.

\bibitem[Zou et~al.(2024)Zou, Yu, Guo, Li, Liang, Cao, and Zhang]{triplaneGaussian}
Zi-Xin Zou, Zhipeng Yu, Yuan-Chen Guo, Yangguang Li, Ding Liang, Yan-Pei Cao, and Song-Hai Zhang.
\newblock Triplane meets gaussian splatting: Fast and generalizable single-view 3d reconstruction with transformers.
\newblock In \emph{Proceedings of the IEEE/CVF Conference on Computer Vision and Pattern Recognition}, pages 10324--10335, 2024.

\end{thebibliography}
}

\end{document}